\documentclass[letterpaper]{article} 
\usepackage[]{aaai2027}  
\usepackage[hyphens]{url}  
\usepackage{graphicx} 
\usepackage{natbib}  
\usepackage{caption} 
\usepackage{algorithm}
\usepackage{algorithmic}

\usepackage{newfloat}
\usepackage{listings}
\DeclareCaptionStyle{ruled}{labelfont=normalfont,labelsep=colon,strut=off} 
\floatstyle{ruled}
\newfloat{listing}{tb}{lst}{}
\floatname{listing}{Listing}

\usepackage{booktabs}

\usepackage[utf8]{inputenc} 
\usepackage[T1]{fontenc}    
\usepackage[table]{xcolor}   

\usepackage{amsmath}
\usepackage{amsfonts}       
\usepackage{amssymb}        
\usepackage{booktabs}       
\usepackage{tabularx}
\usepackage{xcolor}
\usepackage{multirow}
\usepackage{adjustbox}
\usepackage{siunitx}
\usepackage{graphicx}       
\usepackage{svg}
\usepackage{pifont}
\usepackage{nicefrac}       
\usepackage{enumitem}
\usepackage{microtype}      
\usepackage[most]{tcolorbox}

\title{RuleMem: Active Rule Memory for Long-Term Conversational Agents}

\author{
    Xingyuan Zeng\textsuperscript{\rm 1},
    Zuohan Wu\textsuperscript{\rm 2},
    Yue Wang\textsuperscript{\rm 3},
    Chen Zhang\textsuperscript{\rm 4},
    Quanming Yao\textsuperscript{\rm 5},
    Wei Liu\textsuperscript{\rm 1},
    Jiuke Wang\textsuperscript{\rm 1},
    Libin Zheng\textsuperscript{\rm 1}\corresponding,
    Jian Yin\textsuperscript{\rm 1}
}

\affiliations{
    \textsuperscript{\rm 1}Sun Yat-sen University, Zhuhai, China\\
    \textsuperscript{\rm 2}The Hong Kong University of Science and Technology (Guangzhou), Guangzhou, China\\
    \textsuperscript{\rm 3}Shenzhen Institute of Computing Sciences, Shenzhen, China\\
    \textsuperscript{\rm 4}The Hong Kong Polytechnic University, Hong Kong, China\\
    \textsuperscript{\rm 5}Tsinghua University, State Key Laboratory of Space Network and Communications, Beijing National Research Center for Information Science and Technology, Beijing, China\\
    zengxy96@mail2.sysu.edu.cn,
    zh.wu@connect.hkust-gz.edu.cn,
    yuewang@sics.ac.cn,
    jason-c.zhang@polyu.edu.hk,
    qyaoaa@tsinghua.edu.cn,
    liuw259@mail.sysu.edu.cn,
    wangjk57@mail.sysu.edu.cn,
    zhenglb6@mail.sysu.edu.cn,
    issjyin@mail.sysu.edu.cn
}

\begin{document}

\maketitle
\begin{abstract}
Question answering agents in long-term conversations must reason over massive, temporally dispersed dialogue histories. However, existing memory mechanisms primarily treat past information as \textit{passively} stored facts, leading to semantic gaps and unreliable reasoning.
To address this limitation, we propose RuleMem, a rule-based memory framework that induces reusable logical rules from historical interactions to \textit{actively} guide both evidence retrieval and reasoning. Specifically, RuleMem constructs natural-language Horn clauses from conversations and validates them via a Rule Perplexity Consistency (RPC) mechanism. These induced rules enable the retrieval of semantically distant evidence while providing an explicit logical structure for answer generation.
We conducted a comprehensive evaluation of RuleMem on two long-term conversational benchmarks, LoCoMo and LongMemEval\_s*. In a rigorous comparison against 14 baselines on LoCoMo, RuleMem achieved the highest accuracy, exceeding the baseline average by 27.47 points (a 54.3\% relative improvement).
\end{abstract}


\section{Introduction}
\label{sec:intro}

Question answering (QA) over long-term conversations~\cite{locomo} requires agents to reason across massive, unstructured, and temporally dispersed dialogue histories. To overcome the limitations of finite context windows and the difficulties in updating the parametric memory~\cite{param} of large language models (LLMs), external memory modules have become standard components of contemporary agents~\cite{agentmemorysurvey}. These modules enable agents to store past interactions, maintain persistent user or task-related information, and retrieve relevant memories during inference~\cite{fromragtomemory}.

However, QA over long-term conversations faces \textbf{two core challenges: evidence retrieval across semantic gaps}~\cite{nguyen2010bridging} and \textbf{reliable reasoning over the retrieved evidence}~\cite{reasoningonfact}. For example, given the question ``\textit{Why was Alice absent from the meeting?}'', the relevant historical evidence might never \underline{explicitly} mention the word ``\textit{absent}.'' Instead, the system should {reason} based on the record ``\textit{Alice booked a vacation}'' for the final answer. 

Existing memory designs remain superficial and do not fully address both challenges simultaneously. \textbf{We categorize existing agent memory paradigms into two levels of abstraction}: {fact memorization} and {fact organization}.
\begin{figure*}
    \centering
    \includegraphics[width=0.85\linewidth]{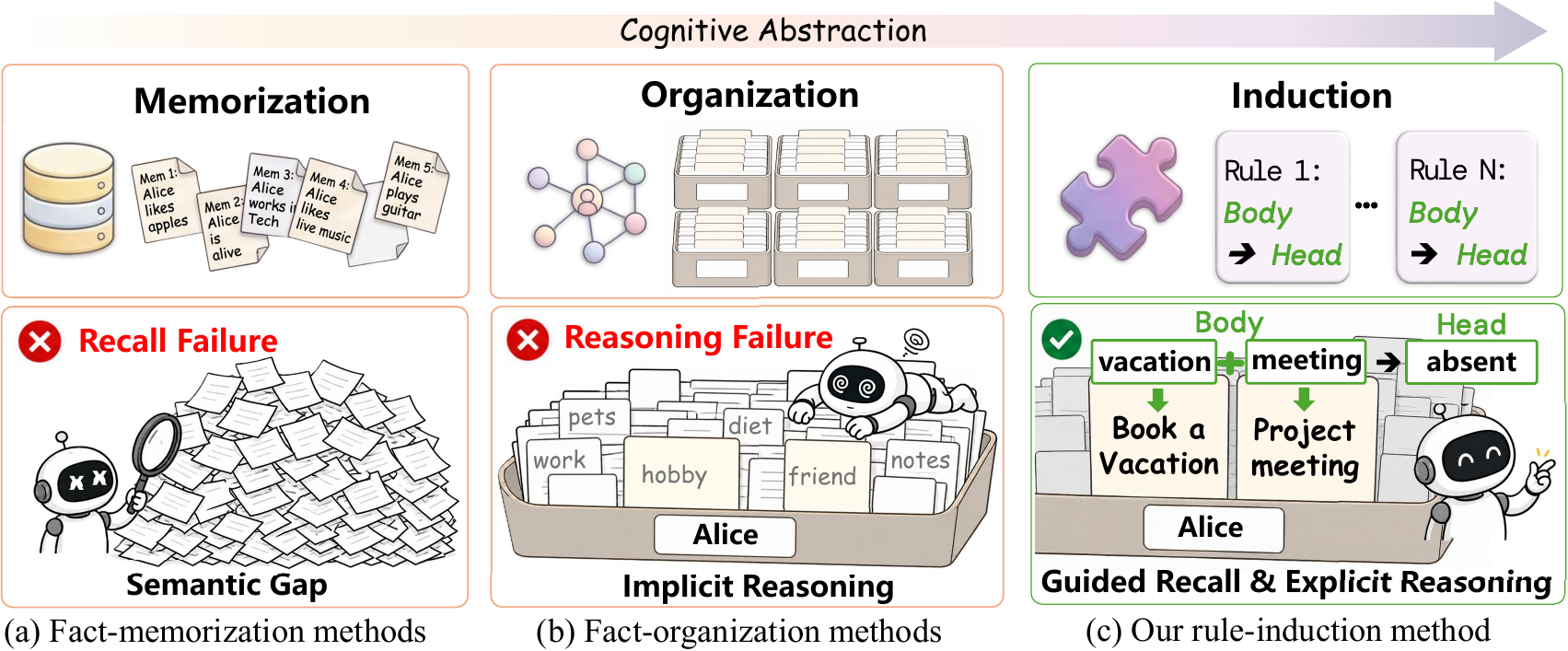}
    \caption{Comparison of three paradigms in agent memory. \textbf{Fact memorization} relies on surface similarity to retrieve isolated snippets. \textbf{Fact organization} builds instance-level structures (e.g., Zettelkasten and knowledge graphs) but introduces noise during complex reasoning. Our proposed \textbf{rule induction} extracts reusable rules from specific facts, bridging semantic gaps for implicit evidence retrieval and providing a logical skeleton for explicit reasoning.}
    \label{fig:intro_paradigms}
\end{figure*}
\textbf{The first level, fact memorization}, focuses on the direct storage and similarity-based retrieval of dialogue snippets or instance-level facts (as exemplified by systems including MemGPT~\cite{packer2023memgpt}, Mem0~\cite{mem0}, and LangMem.)
These methods depend heavily on lexical or shallow semantic overlaps between the current query and stored memories. Consequently, they frequently suffer from recall failures due to underlying semantic gaps.
\textbf{The second level, fact organization}, goes beyond isolated memory entries by introducing structured representations and compressed summaries. For instance, Zep~\cite{zep} and Mem0$^\mathrm{g}$~\cite{mem0} structure dialogue histories into temporal knowledge graphs to establish structural connections among facts; meanwhile, A-MEM~\cite{Amem} and MemInsight~\cite{meminsight} construct Zettelkasten-style~\cite{Luhmann1981} semantic networks and feature-annotated hierarchical summaries, respectively, to correlate and summarize fragmented snippets.
Although these structural connections facilitate the traversal of related facts, they remain confined to the instance level. As the memory scale expands, dense instance connections inevitably introduce query-irrelevant noise. Consequently, LLMs have to infer on their own which facts serve as valid premises and what conclusions they can support. When confronted with lengthy, loosely organized contexts, this deductive reasoning process frequently results in broken logic chains~\cite{Irrelevant,longintput} or hallucinations~\cite{lossinmiddle}, a phenomenon we term reasoning failure.

Observing the limitations above, \textbf{we propose memory exploitation on a higher level of abstraction, rule induction}. 
This idea is drawn from human cognition: memory does not merely {record} or {organize} fragments; rather, it {induces rules from specific observations}~\cite{erickson1998rules}.
After observing that travel plans often lead to meeting absences, humans induce an implicit, reusable rule from specific observations: \textit{``If a person has a recent travel plan, then they may be unable to attend a scheduled event.''}
When faced with a new question about ``absence'', this rule not only indicates what clues should be recalled but also provides a logical skeleton for interpreting the evidence.
Although recent works in sequential decision-making have begun to explore learning workflows~\cite{Workflow,ExpeL, shinn2024reflexion}, the abstraction and learning of inductive rules to assist memory reasoning in QA tasks remain underexplored.

Inspired by this, we propose the \textbf{RuleMem} framework, a rule memory framework for reasoning in question-answering agents.
Unlike existing methods where memory \textit{passively} serves as a recalled object, we propose a novel format as rule memory: \textbf{it induces reusable logical rules from historical conversations to \textit{actively} and \textit{explicitly} guide evidence retrieval and logical deduction for subsequent queries.}
Specifically, we formulate the rule memory using natural-language Horn clauses~\cite{horn1951logic}, combining the structural rigor of formal logic with the semantic flexibility of natural language.
At query time, these rules serve two primary roles. 
First, they guide retrieval. By using rule antecedents as retrieval cues, the model can identify underlying evidence that is semantically distant yet logically relevant, thereby reducing retrieval failures.
Second, they support reasoning. The matched rules serve as explicit major premises~\cite{JohnsonLaird2008}, providing rigorous logical support to connect scattered facts with the final conclusion, thereby mitigating reasoning failures.




Since directly prompting LLMs to induce rules often yields over-generalized or hallucinated results~\cite{hallucination, hallucination2}, RuleMem introduces a \textbf{Rule Perplexity Consistency (RPC)} filtering mechanism.
By measuring reductions in conditional perplexity~\cite{Probabilistic} to evaluate the consistency between internal priors and external factual evidence, RPC effectively filters out unreliable rules to ensure high-quality rule memory. 
The main contributions of this paper are summarized as follows:

\begin{itemize}
    \item \textbf{We propose RuleMem,} a rule memory framework that turns memory from a passively recalled object into an active guide for evidence retrieval and logical deduction.
    \item \textbf{We introduce RPC to verify the quality of induced rules.} RPC uses reductions in conditional perplexity to approximate rule support in a continuous semantic space, jointly evaluating model self-consistency and external factual consistency.

    \item \textbf{We evaluate RuleMem on multiple long-term conversation QA benchmarks.} Experimental results show that RuleMem alleviates recall and reasoning failures inherent in existing methods, outperforming mainstream memory baselines.
\end{itemize}

\section{Related Work}

\noindent\textbf{Memory Mechanisms in LLM Agents.} Memory equips LLM agents to retain past interactions, broadly categorized into factual and experiential forms~\cite{agentmemorysurvey}. 
Factual memory stores declarative knowledge, evolving from flat episodic streams to OS-inspired read-write architectures~\cite{packer2023memgpt, retllm, mem0}. To prevent context fragmentation, recent works organize isolated facts into complex topologies: multi-level user profiles~\cite{Zhong2024MemoryBank}, relational graphs~\cite{zep, Amem, HGMEM}, and chronological timelines~\cite{meminsight, timelinememory}. Beyond static storage, newer models introduce cognitive lifecycles for reflective consolidation and dynamic updates~\cite{RGMem, MemoryR1}.
Experiential memory records task trajectories and executable actions for self-improvement, utilizing text-based reflections~\cite{shinn2024reflexion, ExpeL} or procedural skill libraries~\cite{wang2023voyager, Workflow}. However, these action-level traces are tightly coupled to specific environments, limiting their transferability to conversational QA. Such scenarios require synthesizing generalizable knowledge rather than merely replaying past execution steps.

\noindent\textbf{Retrieval-Augmented Generation (RAG).} RAG grounds LLM outputs in external knowledge to reduce hallucinations~\cite{rag-survey, rag-sur2}. Standard pipelines retrieve flat textual chunks~\cite{guu2020realm, lewis2020rag, borgeaud2022retro}, while advanced structures use hierarchical trees to capture broader contexts~\cite{sarthi2024raptor}. To handle multi-hop relationships, graph-based RAG leverages network topologies by extracting relational paths and subgraphs~\cite{gutierrez2024hipporag, guo2024lightrag, he2024gretriever, li2024dalk}, or generating community-level summaries~\cite{edge2024graphrag, DA-RAG, communityRAG, ArchRAG}. 
Paralleling these structural improvements, agentic paradigms shift RAG from static fetching to autonomous exploration. By employing multi-step reasoning~\cite{yao2022react}, agents adaptively critique their own retrieval steps~\cite{asai2024selfrag}. This autonomy naturally extends to graphs, where agents iteratively navigate and dynamically evaluate retrieval trajectories~\cite{sun2024tog, yuan2025metakgrag}. 
While expanding factual coverage, existing pipelines predominantly supply concrete facts as logical ``minor premises,'' lacking the abstract ``major premises'' necessary for rigorous deductive reasoning~\cite{JohnsonLaird2008}. To resolve this, RuleMem extracts generalized logical rules from past conversations to serve as explicit major premises, systematically guiding the inferential process.

\section{The RuleMem Framework}\label{sec:method}

\begin{figure*}
    \centering
    \includegraphics[width=0.85\linewidth]{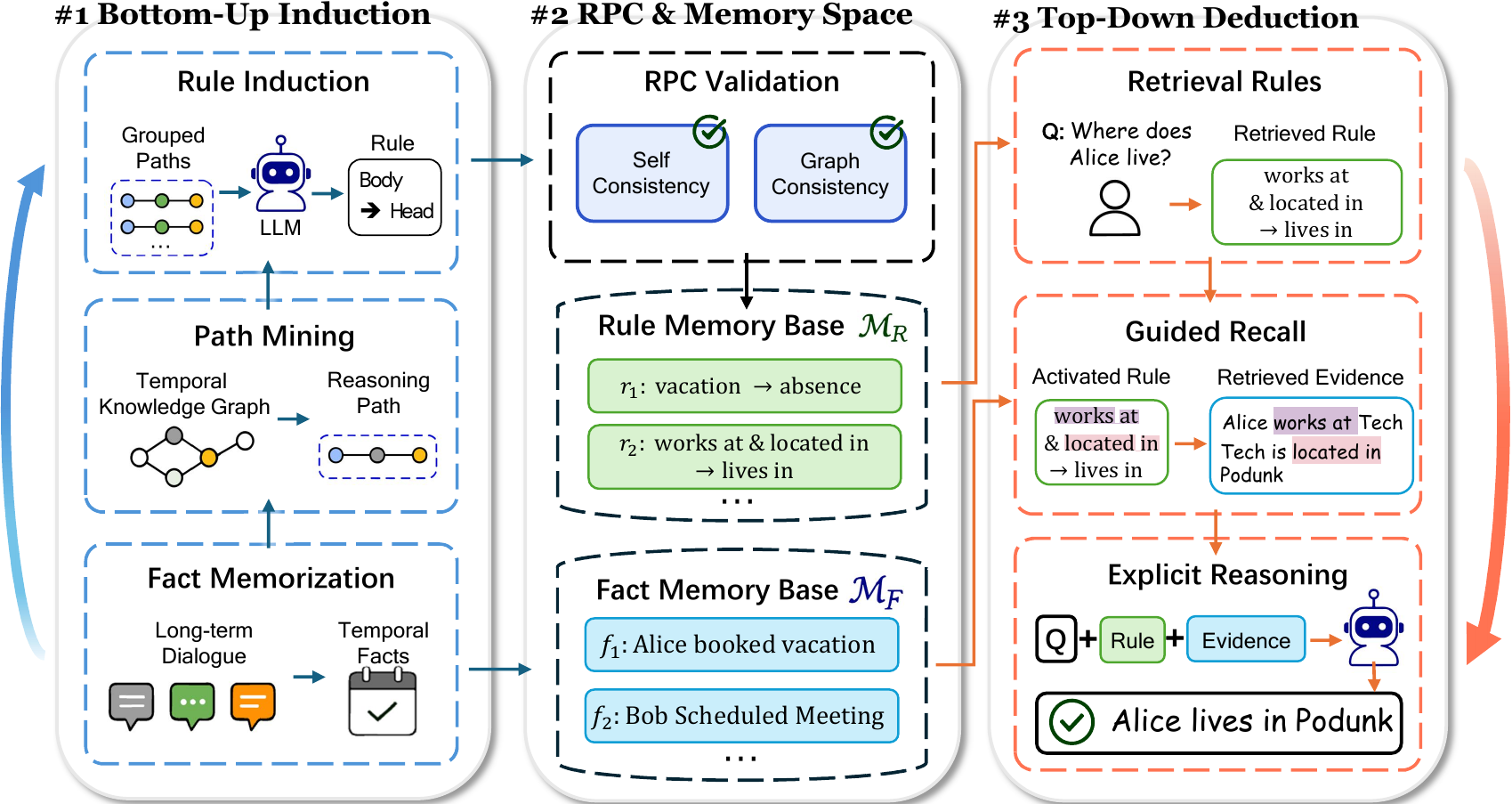}
    \caption{The RuleMem framework. The system operates in three main stages: 1) \textbf{Bottom-up rule memory construction}: extracts temporal facts from dialogues to form a fact memory base (\(\mathcal{M}_F\)), mines reasoning paths, and induces them into abstract natural-language rules stored in a rule memory base (\(\mathcal{M}_R\)). 2) \textbf{Rule validation via RPC}: filters and refines the induced rules by combining internal language priors with external factual consistency. 3) \textbf{Top-down rule-guided QA}: when facing a new question, the system activates relevant rule templates, guides the recall of supporting facts, and generates explicit reasoning answers using both the rule scaffold and factual evidence.}
    \label{fig:framework}
\end{figure*}

\textbf{Overview: Bottom-Up Induction $\rightarrow$ Rule Validation $\rightarrow$ Top-Down Deduction}. RuleMem organizes long-term conversational memory as a closed loop of bottom-up induction, RPC validation, and top-down deduction. As shown in Figure~\ref{fig:framework}, RuleMem maintains two memory spaces: a fact memory base \(\mathcal{M}_F\) and a rule memory base \(\mathcal{M}_R\). The former stores temporal facts extracted from dialogues, while the latter stores natural-language rules induced from concrete reasoning traces. During construction, RuleMem moves from fact extraction to path sampling and rule induction, lifting concrete facts into abstract regularities and validating them with Rule Perplexity Consistency (RPC). During QA, RuleMem uses these rules in reverse: it activates relevant rules, recalls supporting facts, and generates answers under rule-guided constraints.

\subsection{Bottom-Up Rule Memory Construction}\label{sec:bottom-up-construction}

\textbf{Fact Memorization.} The fact memorization stage converts unstructured, multi-turn dialogue content into retrievable memory units. RuleMem first parses long-term dialogues and extracts subject entities, relations, object entities, and timestamps as temporal quadruples \(\mathcal{F} = \{(e_s, r, e_o, t)\}\). These quadruples constitute the fact memory base \(\mathcal{M}_F\). It preserves concrete facts that can be reused by later path mining and evidence recall, providing a traceable experiential basis for rule induction.

\textbf{Reasoning Path Mining.} RuleMem organizes \(\mathcal{F}\) in a graph-structured layout for path sampling. Over this structure, the system first samples candidate paths via constrained random walks \cite{randomwalk}. An LLM filters out illogical candidate paths and reconstructs valid ones from the fact memory base along with their corresponding raw dialogue snippets to distill coherent reasoning paths \(\mathcal{P}_c\). These raw paths, which preserve concrete logical relations, directly serve as ungeneralized precursors for downstream rule induction.

\textbf{Rule Induction.} Although reasoning paths reveal connections among multi-hop facts, they remain instance-level memories and are difficult to transfer directly to new questions. The system groups paths with identical or similar relational patterns and uses an LLM to induce shared rules from them. During this process, the LLM distills general causal patterns from complex facts by abstracting away specific entity names (e.g., Alice, Bob, Marathon) and replacing them with typed placeholders such as \texttt{[Person]}, \texttt{[Event]}, or \texttt{[Location]}. Following the core structure of Horn clauses~\cite{horn1951logic}, which consist of a non-empty conjunctive body and a single atomic head, we express each \(r\in\mathcal{M}_R\) as a Horn clause:
\begin{equation}
    r:\;\underbrace{B_{1}\;\wedge\;\cdots\;\wedge\;B_{n}}_{\text{Body }T_A^{(r)}\;,\;n\geq 1}\;\Longrightarrow\;\underbrace{H}_{\text{Head }T_C^{(r)}\;},
\end{equation}
where each atom \(B_i\) or \(H\) is a natural-language relational phrase that may contain type placeholders. 

\subsection{Rule Validation via RPC}\label{sec:rpc}

The rule level provides generalizable reasoning templates, but induction summarizes patterns from finite examples and therefore does not guarantee correctness. As a result, LLM-induced rules may be overgeneralized or weakly grounded in the underlying facts. RPC uses perplexity reduction to test whether a rule is supported by both the model's internal prior and external factual evidence. Given an autoregressive model \(M\) (e.g., Llama-3-8B-Instruct), let \(\ell_M(T \mid C)\) denote the average negative log-likelihood of target text \(T\) conditioned on context \(C\). The internal-consistency signal is defined as:
\begin{equation}
    \Delta_{\text{self}}(r) =
    \ell_M(T_C^{(r)} \mid \varnothing) -
    \ell_M(T_C^{(r)} \mid T_A^{(r)}).
\end{equation}

Relying only on the body and head of the rule may accept rules that are plausible under the language prior but insufficiently grounded in facts. RPC therefore further retrieves related evidence \(T_E^{(r)}\) from the fact memory base and measures whether this evidence further reduces the perplexity of the conclusion given the premise: 
\begin{equation}
    \Delta_{\text{fact}}(r) =
    \ell_M(T_C^{(r)} \mid T_A^{(r)}) -
    \ell_M(T_C^{(r)} \mid T_A^{(r)} \oplus T_E^{(r)}).
\end{equation}
The final RPC confidence score combines the internal-consistency signal and the external factual-consistency signal:
\(
    \operatorname{RPC}(r) =
    \alpha \cdot \sigma(\Delta_{\text{self}}(r)) +
    (1-\alpha) \cdot \sigma(\Delta_{\text{fact}}(r)).
\)
The balancing coefficient \(\alpha \in [0,1]\) controls the relative weight of the two signals, and \(\sigma\) represents the Sigmoid function used to normalize the perplexity reductions into the \((0,1)\) range. Only rules with an RPC score exceeding a threshold \(\tau\) are added to the rule memory base. Empirical tuning and sensitivity of these parameters are discussed in Section~\ref{subsec:hyperparameters_backbone}. 

\subsection{Top-Down Rule-Guided Question Answering}\label{sec:top-down-qa}

\textbf{Retrieval Rules.} During QA, RuleMem does not move directly from the question to factual retrieval. Instead, it first identifies applicable reasoning goals at the rule level. Given a new question \(Q\), the system activates candidate rules by matching the question with the rule head $T_C^{(r)}$ in the rule memory base: 

\begin{equation}
    \mathcal{R}_{\text{active}} =
    \operatorname*{TopN}_{r \in \mathcal{M}_R}
    \operatorname{Cos}\left(\mathbf{e}(Q), \mathbf{e}(T_C^{(r)})\right).
\end{equation}

This matching maps the question into an abstract conclusion space, allowing the system to locate a relevant rule before relying on surface similarity between the question and facts.

\textbf{Guided Recall.} The activated rules first expose their premise conditions, i.e., their body $T_A^{(r)}$, as retrieval cues to recall a set of candidate facts $\mathcal{E}_{\text{cand}}^{(r)}$ from the fact memory base:
\begin{equation}
    \mathcal{E}_{\text{cand}}^{(r)} =
    \operatorname*{TopK}_{f \in \mathcal{M}_F}
    \operatorname{Cos}\left(\mathbf{e}(T_A^{(r)}), \mathbf{e}(f)\right).
\end{equation}
At this stage, the abstract typed variables lack formal variable binding and unification, which risks mismatching evidence (e.g., retrieving facts about the wrong entity). To accurately instantiate the rule in the new context, RuleMem introduces an LLM-based filtering step. By referencing the specific entities in the question $Q$ and the typed variable constraints in the rule body, the LLM acts as a semantic unification operator. It evaluates the recalled facts, discarding those that violate the variable bindings and typing constraints, and retains only the accurately grounded evidence as the final $\mathcal{E}_{\text{guided}}^{(r)}$.

\textbf{Explicit Reasoning.} Finally, RuleMem organizes the question, activated rules, and recalled evidence into a structured prompt, allowing the generation model to observe both the reasoning scaffold and the factual support. At this stage, the rules provide a reasoning template analogous to deductive application in logic, while the evidence provides episodic details analogous to recalled experience in cognition, mitigating reasoning failure. \textbf{The complete prompt templates are detailed in the appendix.}

\begin{table*}[!t]
    \centering
    \small
    \setlength{\tabcolsep}{4.5pt}
    \renewcommand{\arraystretch}{1.15}

    \begin{adjustbox}{width=\textwidth}
    \begin{tabular}{l
                    S[table-format=2.2] S[table-format=2.2] S[table-format=2.2]
                    S[table-format=2.2] S[table-format=2.2] S[table-format=2.2]
                    S[table-format=2.2] S[table-format=2.2] S[table-format=2.2]
                    S[table-format=2.2] S[table-format=2.2] S[table-format=2.2]
                    S[table-format=2.2] S[table-format=2.2] S[table-format=2.2]}
      \toprule
      & \multicolumn{3}{c}{Single-hop}
      & \multicolumn{3}{c}{Multi-hop}
      & \multicolumn{3}{c}{Open-domain}
      & \multicolumn{3}{c}{Temporal}
      & \multicolumn{3}{c}{Avg.} \\
      \cmidrule(lr){2-4} \cmidrule(lr){5-7} \cmidrule(lr){8-10} \cmidrule(lr){11-13} \cmidrule(lr){14-16}
      Method
      & {F1} & {BLEU} & {Acc}
      & {F1} & {BLEU} & {Acc}
      & {F1} & {BLEU} & {Acc}
      & {F1} & {BLEU} & {Acc}
      & {F1} & {BLEU} & {Acc} \\
      \midrule
  
  
  
  
      \rowcolor{gray!15} 
      \multicolumn{16}{l}{\fontsize{8pt}{10pt}\selectfont \textit{Fact Memorization}} \\

      Mem0
      & \underline{39.06} & 31.97 & 59.64
      & \underline{25.73} & 17.54 & 42.98
      & 16.74 & 11.58 & 37.50
      & \textbf{53.99} & 44.64 & 57.58
      & \textbf{33.88} & 26.43 & 49.43 \\

      Letta
      & 29.93 & 19.33 & \underline{85.90}
      & 22.55 & 17.18 & 65.96
      & 15.21 & 12.32 & 51.04
      & 6.99 & 5.54 & 39.56
      & 18.67 & 13.59 & 60.62 \\

      LangMem
      & 13.67 & 8.48 & 46.85
      & 12.00 & 9.55 & 50.35
      & 8.59 & 6.12 & 47.92
      & 5.32 & 4.20 & 11.84
      & 9.90 & 7.09 & 39.24 \\
      \rowcolor{gray!15} 
      \multicolumn{16}{l}{\fontsize{8pt}{10pt}\selectfont \textit{Fact Organization}} \\

      A-MEM
      & 36.22 & 30.01 & 58.31
      & 22.07 & 16.98 & 46.10
      & 12.86 & 10.80 & 30.21
      & 37.77 & 33.06 & 44.24
      & 27.23 & 22.71 & 44.72 \\

      Mem0\(^\mathrm{g}\)
      & 25.81 & 31.42 & 43.56
      & 21.95 & \underline{32.63} & 49.73
      & 11.95 & 11.41 & 33.74
      & 35.05 & 43.66 & 44.30
      & 23.69 & 29.78 & 42.83 \\

      MemoryBank
      & 18.54 & 11.50 & 55.29
      & 13.11 & 10.61 & 36.88
      & 9.83 & 7.87 & 45.83
      & 17.65 & 12.01 & 43.61
      & 14.78 & 10.50 & 45.40 \\

      MemInsight
      & \textbf{39.75} & 33.97 & 63.21
      & 21.97 & 15.00 & 49.29
      & 15.70 & 13.35 & 40.62
      & 6.99 & 5.98 & 44.20
      & 21.10 & 17.08 & 49.33 \\

      Zep
      & 18.93 & 10.25 & 81.07
      & 13.47 & 9.00 & 76.24
      & 8.58 & 4.86 & 66.67
      & 21.33 & 14.25 & \textbf{66.67}
      & 15.58 & 9.59 & \underline{72.66} \\

      SCM
      & 24.34 & 14.34 & 81.33
      & 15.57 & 11.72 & 65.60
      & 10.99 & 6.72 & 60.42
      & 6.06 & 5.01 & 43.93
      & 14.24 & 9.45 & 62.82 \\

      \rowcolor{gray!15} 
      \multicolumn{16}{l}{\fontsize{8pt}{10pt}\selectfont \textit{RAG Methods}} \\


      BM25
      & 18.86 & 11.25 & 57.91
      & 9.62 & 8.02 & 31.56
      & 8.44 & 5.09 & 42.71
      & 3.89 & 3.52 & 23.36
      & 10.20 & 6.97 & 38.89 \\

      ReAct
      & 30.26 & 26.67 & 50.42
      & 15.10 & 10.39 & 31.21
      & 9.17 & 7.36 & 29.17
      & 14.02 & 9.75 & 26.79
      & 17.14 & 13.54 & 34.40 \\

      MetaKGRAG
      & 17.85 & 16.12 & 42.09
      & 12.10 & 9.84 & 33.29
      & 7.31 & 6.62 & 28.31
      & 13.48 & 11.63 & 33.09
      & 12.68 & 11.05 & 34.19 \\

      LightRAG
      & 15.82 & 7.53 & \textbf{87.87}
      & 9.88 & 6.09 & \underline{79.79}
      & 6.36 & 4.05 & 46.88
      & 10.27 & 6.09 & 43.93
      & 10.58 & 5.94 & 64.62 \\

      GraphRAG
      & 24.61 & 15.43 & 85.73
      & 20.10 & 16.14 & 73.05
      & 12.49 & 10.28 & 67.71
      & 15.46 & 10.80 & 49.53
      & 18.17 & 13.16 & 69.01 \\

      \midrule
      \rowcolor{blue!10}
      \textbf{Ours}
      & 38.02 & \textbf{42.12} & 85.73
      & \textbf{26.97} & \textbf{38.74} & \textbf{82.43}
      & \underline{17.86} & \underline{19.00} & \textbf{78.37}
      & \underline{49.00} & \textbf{47.74} & \underline{65.66}
      & \underline{32.96} & \textbf{36.90} & \textbf{78.05} \\
      \midrule
      \rowcolor{gray!15} 
      \multicolumn{16}{l}{\fontsize{8pt}{10pt}\selectfont \textit{Ablation Variants}} \\

      w/o RPC
      & 29.23 & \underline{35.69} & 72.50
      & 13.67 & 20.42 & 55.89
      & \textbf{18.98} & \textbf{19.77} & \underline{78.26}
      & 37.40 & \underline{46.12} & 56.94
      & 24.82 & \underline{30.50} & 65.90 \\

      w/o Rule+RPC
      & 11.05 & 20.26 & 51.87
      & 9.29 & 14.09 & 38.24
      & 5.33 & 9.23 & 34.07
      & 14.92 & 22.32 & 49.52
      & 10.15 & 16.48 & 43.43 \\
  
      \bottomrule
    \end{tabular}
    \end{adjustbox}
        \caption{Performance comparison on the LoCoMo benchmark (best in \textbf{bold}, second best \underline{underlined}), across four question types: \textit{Single-hop} (single evidence), \textit{Multi-hop} (cross-snippet reasoning), \textit{Open-domain} (dialogue with external knowledge), and \textit{Temporal} (time-aware reasoning).}
    \label{tab:results}
        
  \end{table*}

\begin{figure*}[!ht]
  \centering
  \includegraphics[width=\textwidth]{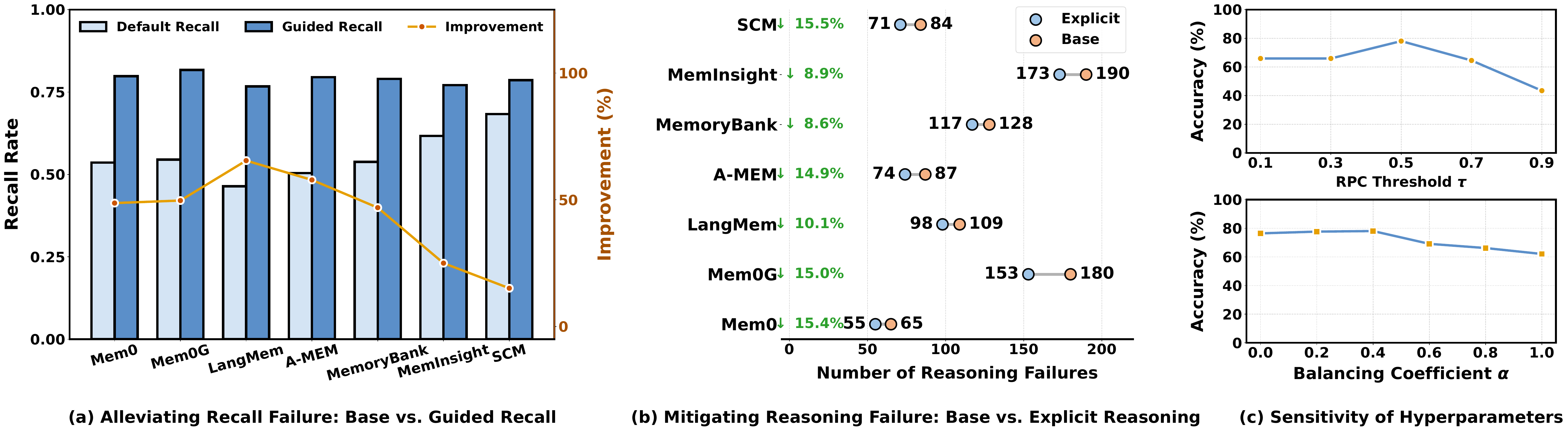}
  \caption{Analysis of recall failure, reasoning failure, and RPC hyperparameter sensitivity. \textbf{(a)} Recall rates obtained with Default Recall and Guided Recall across memory baselines; the orange line reports the relative improvement from guided retrieval. \textbf{(b)} Numbers of reasoning failures under default generation and Explicit Reasoning; the arrows report the relative reduction for each baseline. \textbf{(c)} Accuracy sensitivity to the RPC admission threshold $\tau$ (top) and the signal balancing coefficient $\alpha$ (bottom).}
  \label{fig:analysis_all}
\end{figure*}
\section{Experiments}
  \label{sec:experiments}
In this section, we conduct experiments designed to answer the following research questions (RQs):

    \noindent$\bullet$ \textbf{RQ1:} How does RuleMem perform compared to mainstream memory-augmented and reasoning baselines on long-term conversation and extended context benchmarks?

    \noindent$\bullet$ \textbf{RQ2:} How do the proposed Guided Recall and Explicit Reasoning mechanisms effectively alleviate the specific challenges of recall failure and reasoning failure?

    \noindent$\bullet$ \textbf{RQ3:} How robust is RuleMem regarding base models and hyperparameters?

  \subsection{Experimental Setup}
  \label{subsec:setup}
  
  \textbf{\textit{Datasets and Evaluation Metrics.}} Following the common evaluation protocols in recent memory-augmented agent research, we evaluate our proposed RuleMem framework on two widely used benchmark datasets:
  1) \textbf{LoCoMo}~\cite{locomo}: A comprehensive long-term conversation memory dataset, comprising 5,882 dialogue turns, and 1,986 questions in total.
  2) \textbf{LongMemEval\_s*}: A subset of the \textbf{MemoryAgentBench}~\cite{memoryagentbench,LongMemEval} dataset, reformulated into 5 long dialogue sequences with approximately 1.82M tokens, and 300 questions in total (results are reported in the appendix).
  For LoCoMo, \textbf{our evaluation scripts and metrics are strictly consistent} with the Mem0~\cite{mem0} repository, utilizing the F1 score, BLEU, and Accuracy to comprehensively evaluate generation quality and accuracy; similarly, for LongMemEval\_s*, we strictly employ the original evaluation scripts and metrics from MemoryAgentBench~\cite{memoryagentbench}, reporting task-specific accuracy (detailed metrics are provided in the appendix).

\textit{\textbf{Baselines.}}
We compare our method with 14 baselines categorized into three groups. The fact memorization group includes \textbf{Mem0}~\cite{mem0}, \textbf{LangMem}, and \textbf{Letta}~\cite{packer2023memgpt}. The fact organization group consists of \textbf{A-MEM}~\cite{Amem}, \textbf{Mem0$^\mathrm{g}$}~\cite{mem0}, \textbf{MemoryBank}~\cite{Zhong2024MemoryBank}, \textbf{MemInsight}~\cite{meminsight}, \textbf{Zep}~\cite{zep}, and \textbf{SCM}~\cite{SCM}. Lastly, the RAG group comprises \textbf{BM25}, \textbf{ReAct}~\cite{yao2022react}, \textbf{MetaKGRAG}~\cite{yuan2025metakgrag}, \textbf{LightRAG}~\cite{guo2024lightrag}, and \textbf{GraphRAG}~\cite{edge2024graphrag}. Together, they cover similarity-based fact retrieval, structured memory organization, and text- or graph-based retrieval, enabling a comprehensive evaluation across memory and retrieval paradigms. Full model descriptions and configurations are provided in the appendix.

  \textbf{\textit{Implementation Details.}} In our experiments, we employ \texttt{gpt-4o-mini} as the backbone LLM for the abstraction of memory rules and the final explicit reasoning phase. For the vector database, we utilize \texttt{ChromaDB} with its default embedding model, \texttt{all-MiniLM-L6-v2}, to store embeddings of facts and rules. 
  Through experiments, the RPC admission threshold $\tau$ is empirically set to $0.5$, and the balancing coefficient $\alpha$ is set to $0.4$. Unless otherwise specified, all baseline methods are evaluated using their official repository implementations with default parameters.
  The complete prompt templates and extended implementation details can be found in the appendix.

\subsection{Main Results (RQ1)} 
\label{subsec:main_results}

As shown in Table~\ref{tab:results}, RuleMem demonstrates strong overall performance across the four question types, with an average BLEU of 36.90 and accuracy of 78.05. This highlights the effectiveness of the rule memory paradigm for complex dialogue reasoning.

Specifically, in Multi-hop and Open-domain tasks that require deeper logical deduction, RuleMem shows substantial improvements. For instance, its accuracy in multi-hop scenarios reaches 82.43, outperforming the best baseline (79.79). Pure fact memorization (e.g., Mem0, Letta) and traditional RAG pipelines (e.g., ReAct, GraphRAG) typically retrieve concrete facts as ``minor premises'' but lack the abstract ``major premises'' necessary for deductive reasoning~\cite{JohnsonLaird2008}. Fact organization methods (e.g., A-MEM, Zep) also struggle to maintain logical chains across long contexts. In contrast, RuleMem extracts abstract rules to serve as major premises, facilitating more rigorous logical deduction.
For Single-hop questions relying on a single pieces of evidence, RuleMem achieves the highest generation quality (BLEU: 42.12) and maintains competitive accuracy. We attribute this to the Guided Recall mechanism. Retrieved abstract rules act as precise cues, helping the model anchor relevant facts within long contexts. This prevents the loss of specific details and improves the structural clarity of the generated responses.
Due to space constraints, extended results on the LongMemEval\_s* dataset are provided in the appendix.

\begin{table*}[!ht]
    \centering
    \small
    \renewcommand{\arraystretch}{1.15}
    \begin{tabularx}{\textwidth}{>{\raggedright\arraybackslash}p{0.25\textwidth}X}
    \toprule
    \textbf{Component} & \textbf{Content} \\
    \midrule
    
    Question \& Gold answer
    &
    \textbf{Question:} ``What did Caroline realize after her charity race?'' \newline
    \textbf{Gold answer:} ``Self-care is important.'' \\

    \midrule
    
    Conversation excerpt
    &
    Melanie: ``Last month I \textbf{got hurt} and had to take a break from pottery''
    
    
    \ldots
    
    Melanie: ``Thanks, Caroline! The event was really thought-provoking. I'm starting to realize that \textbf{self-care is really important}.''
    
    Caroline: ``\textbf{I totally agree with you.} It's a journey for me.'' \\
    
    \bottomrule
    
    Mem0 retrieved memory excerpts
    &
    ``Found the \textbf{charity race} rewarding.'' \emph{(high vector similarity to the event)}
    
    ``Ran a charity race for mental health last Saturday.'' \emph{(event-level match)}
    
    ``Melanie found the experience scary.'' \emph{(experience-level match)}

    \ldots \\
    
    \midrule
    
    Mem0 response
    &
    ``Melanie found it rewarding.'' \\
    
    \bottomrule
    
    Mem0$^\mathrm{g}$ retrieved graph memory excerpts
    &
    $(\texttt{melanie}, \texttt{found\_in}, \texttt{inspiration})$ \emph{(loosely related to realization/reflection)}
    
    $(\texttt{melanie}, \texttt{experiences}, \texttt{setback})$ \emph{(loosely related to difficulty/injury)}

    $(\texttt{melanie}, \texttt{reminds\_of}, \texttt{family\_love})$ \emph{(related to family but not self-care)}

    \ldots \\
    
    \midrule
    
    Mem0$^\mathrm{g}$ response
    &
    ``Melanie found the race rewarding.'' \\
    
    \bottomrule
    
    RuleMem activated rule
    &
    [Person] participates in [Physical Event] $\wedge$ [Person] is injured during [Physical Event] $\rightarrow$ [Person] may realize that self-care is important. \\

    \midrule
    
    RuleMem recalled memory excerpts
    &
    ``Melanie participated in the charity race.'' \emph{(matches [Person] participates in [Physical Event])}
    
    ``Melanie was injured during the charity race.'' \emph{(matches [Person] is injured during [Physical Event])} \\

    \midrule
    
    RuleMem response
    &
    ``Melanie realized that self-care is important.'' \\
    
    \bottomrule
    \end{tabularx}

    \caption{Case study of rule-guided recall and explicit reasoning. The conversation and retrieved memories are abbreviated excerpts. Bold text highlights key reasoning-relevant contents, while parentheses indicate why each memory was retrieved or how it relates to the query.}
\label{tab:case_study_rulemem}
\end{table*}

\subsection{In-depth Analysis (RQ2)}
To understand the sources of RuleMem's performance gains, we analyze its effectiveness in mitigating two primary challenges: recall failure and reasoning failure. We also conduct an ablation study to isolate component contributions, followed by case studies and error analysis.

\textit{\textbf{Alleviating recall failure.}}
Standard retrieval often fails when evidence is semantically distant from the query. RuleMem addresses this via \textit{Guided Recall}, which uses the body of activated rules as supplementary search constraints. Using LoCoMo's~\cite{locomo} manually annotated supporting facts, Fig.~\ref{fig:analysis_all}(a) shows that Guided Recall consistently improves recall rates across all frameworks, raising the average recall from 0.56 to 0.79 (+41.1\%). This indicates that rule-conditioned premises successfully broaden the search scope to retrieve logically related facts that lack direct keyword overlap.

\textit{\textbf{Mitigating reasoning failure.}}
Even with correct evidence retrieved, LLMs can fail to deduce answers from long or disorganized contexts. We define \textit{reasoning failure} as cases where facts are successfully retrieved but the final answer remains incorrect. We compare default generation (using only retrieved evidence) with Explicit Reasoning (injecting abstract rules as a logical skeleton). As shown in Fig.~\ref{fig:analysis_all}(b), Explicit Reasoning consistently reduces reasoning failures from an average of 120.4 to 105.9 (-12.0\%), with the largest reduction observed in Mem0$^\mathrm{g}$. This confirms that abstract rules effectively guide multi-step inference and keep the LLM aligned with logical dependencies.

\textit{\textbf{Ablation study.}} \label{subsubsec:ablation}
To evaluate the contribution of each core component, we compare RuleMem against two variants (shown at the bottom of Table~\ref{tab:results}).
  \textbf{w/o Rule+RPC}: Removes rule abstraction and explicit reasoning, reducing the system to a standard fact-organization memory.
   \textbf{w/o RPC}: Commits all induced rules to memory without the perplexity-based reliability check.

Removing rule abstraction (\textbf{w/o Rule+RPC}) causes a severe performance drop, reducing metrics to levels similar to baseline fact-organization methods. This highlights the necessity of high-level rule abstraction for long-term reasoning. Similarly, the absence of RPC filtering (\textbf{w/o RPC}) also degrades performance, as unfiltered, erroneous, or hallucinated rules interfere with the deduction process. The perplexity-based consistency check is therefore essential for ensuring rule reliability.

\subsection{Case Analysis}
\label{sec:case_study}

We examine a successful case and a failure case to illustrate how rule memory affects retrieval and reasoning. These examples complement the aggregate results by revealing both RuleMem's mechanism and its remaining limitation.

\textit{\textbf{Successful Case.}} The question asks what Caroline realized after a charity race, and the gold answer is that self-care is important (Table~\ref{tab:case_study_rulemem}). Mem0 retrieves memories that are topically similar to the question, such as finding the race rewarding or scary, but misses the injury-related evidence and answers that Melanie found the race rewarding. Mem0$^\mathrm{g}$ retrieves a graph neighborhood containing loosely related associations about inspiration and setbacks. However, these instance-level connections neither identify the required premise combination nor explain how an injury can lead to a realization about self-care, so it also produces an incorrect answer.

RuleMem instead activates the rule ``[Person] participates in [Physical Event] $\wedge$ [Person] is injured during [Physical Event] $\rightarrow$ [Person] may realize that self-care is important.'' Its antecedents guide recall toward two supporting facts: Melanie participated in the charity race and was injured during it. RuleMem then uses the activated rule as a major premise and the recalled facts as minor premises, correctly concluding that Melanie realized the importance of self-care. This case illustrates how rule memory guides both evidence retrieval and answer generation.

\textit{\textbf{Failure Case.}} The aggregate analyses above report fewer recall and reasoning failures under the evaluated comparisons, but RuleMem still produces incorrect answers. For instance, consider the following conversation excerpt:
\begin{quote}
    \emph{John:} ``Thank you very much! Will there be some kind of interview required?'' \\
    \emph{James:} ``\textbf{No, this is not necessary.} All you need is to be a friendly and polite person. I'm sure you will succeed!''
\end{quote}
When asked the question, ``Will there be an interview required to volunteer with the organization James volunteered for?'', the gold answer should be ``No,'' as the conversation explicitly states that an interview is not necessary. However, RuleMem retrieves and activates a generalized rule: [Person] is friendly and polite $\rightarrow$ [Person] may pass the interview. Consequently, RuleMem produces an incorrect response: ``Yes, there will be some kind of interview required to volunteer.'' The model over-generalizes based on this activated rule and ignores the specific, contradictory factual evidence present in the dialogue.

  \subsection{Impact of Hyperparameters (RQ3)}
  \label{subsec:hyperparameters_backbone}

  \textit{\textbf{Sensitivity of Hyperparameters.} }We evaluate the RPC mechanism's sensitivity to the admission threshold $\tau$ and the signal balancing coefficient $\alpha$ (Figure~\ref{fig:analysis_all}c).
A low threshold ($\tau \le 0.3$) admits over-generalized rules that misguide reasoning, while an excessively high threshold ($\tau > 0.8$) leads to a sparse rule memory, degrading the system to unguided retrieval.
  An optimal balance at $\tau = 0.5$ ensures both rule reliability and sufficient coverage.
  For $\alpha$, relying solely on the internal language prior ($\alpha=1.0$) causes a significant performance drop.
  The peak performance at $\alpha = 0.4$ indicates that while external factual grounding plays a dominant role in eliminating hallucinations, a moderate internal prior further enhances rule generalizability.
  
\textit{  \textbf{Effect of Base Models.}} We also evaluate the generalizability and robustness of RuleMem across different underlying LLMs, we conduct experiments using three distinct base models: \texttt{gpt-4o-mini}, \texttt{gpt-4o}, and a recent open-weight model \texttt{qwen3-next-80b-a3b-instruct}. RuleMem consistently outperforms all baseline methods across all three base models. For detailed results, please refer to the appendix.
\section{Conclusion}
\label{sec:conclusion}

RuleMem is a framework that transforms agent memory from passive storage into an active tool for evidence retrieval and logical deduction. By learning rules from past interactions, RuleMem improves retrieval of relevant evidence and supports reasoning. Experiments show Guided Recall finds logical evidence, Explicit Reasoning reduces errors, and the RPC mechanism filters unreliable rules. RuleMem demonstrates that memory should actively guide retrieval and reasoning, helping agents learn and deduce from experience.

\bibliography{references,ref}

@inproceedings{locomo,
  author    = {Adyasha Maharana and
               Dong{-}Ho Lee and
               Sergey Tulyakov and
               Mohit Bansal and
               Francesco Barbieri and
               Yuwei Fang},
  title     = {Evaluating Very Long-Term Conversational Memory of {LLM} Agents},
  booktitle = {{ACL} {(1)}},
  pages     = {13851--13870},
  publisher = {Association for Computational Linguistics},
  year      = {2024}
}

@inproceedings{randomwalk,
  author       = {Jiapu Wang and
                  Kai Sun and
                  Linhao Luo and
                  Wei Wei and
                  Yongli Hu and
                  Alan Wee{-}Chung Liew and
                  Shirui Pan and
                  Baocai Yin},
  title        = {Large Language Models-guided Dynamic Adaptation for Temporal Knowledge
                  Graph Reasoning},
  booktitle    = {NeurIPS},
  year         = {2024}
}

@article{memoryagentbench,
  author       = {Yuanzhe Hu and
                  Yu Wang and
                  Julian J. McAuley},
  title        = {Evaluating Memory in {LLM} Agents via Incremental Multi-Turn Interactions},
  journal      = {CoRR},
  volume       = {abs/2507.05257},
  year         = {2025}
}

@Inbook{Luhmann1981,
author="Luhmann, Niklas",
editor="Baier, Horst
and Kepplinger, Hans Mathias
and Reumann, Kurt",
title="Kommunikation mit Zettelk{\"a}sten",
bookTitle="{\"O}ffentliche Meinung und sozialer Wandel / Public Opinion and Social Change",
year="1981",
publisher="VS Verlag f{\"u}r Sozialwissenschaften",
address="Wiesbaden",
pages="222--228",
isbn="978-3-322-87749-9",
doi="10.1007/978-3-322-87749-9_19",
url="https://doi.org/10.1007/978-3-322-87749-9_19"
}

@inproceedings{fromragtomemory,
  author    = {Bernal Jim{\'{e}}nez Guti{\'{e}}rrez and
               Yiheng Shu and
               Weijian Qi and
               Sizhe Zhou and
               Yu Su},
  title     = {From {RAG} to Memory: Non-Parametric Continual Learning for Large
               Language Models},
  booktitle = {{ICML}},
  series    = {Proceedings of Machine Learning Research},
  publisher = {{PMLR} / OpenReview.net},
  year      = {2025}
}

@article{agentmemorysurvey,
  author  = {Yuyang Hu and
             Shichun Liu and
             Yanwei Yue and
             Guibin Zhang and
             Boyang Liu and
             Fangyi Zhu and
             Jiahang Lin and
             Honglin Guo and
             Shihan Dou and
             Zhiheng Xi and
             Senjie Jin and
             Jiejun Tan and
             Yanbin Yin and
             Jiongnan Liu and
             Zeyu Zhang and
             Zhongxiang Sun and
             Yutao Zhu and
             Hao Sun and
             Boci Peng and
             Zhenrong Cheng and
             Xuanbo Fan and
             Jiaxin Guo and
             Xinlei Yu and
             Zhenhong Zhou and
             Zewen Hu and
             Jiahao Huo and
             Junhao Wang and
             Yuwei Niu and
             Yu Wang and
             Zhenfei Yin and
             Xiaobin Hu and
             Yue Liao and
             Qiankun Li and
             Kun Wang and
             Wangchunshu Zhou and
             Yixin Liu and
             Dawei Cheng and
             Qi Zhang and
             Tao Gui and
             Shirui Pan and
             Yan Zhang and
             Philip Torr and
             Zhicheng Dou and
             Ji{-}Rong Wen and
             Xuanjing Huang and
             Yu{-}Gang Jiang and
             Shuicheng Yan},
  title   = {Memory in the Age of {AI} Agents},
  journal = {CoRR},
  volume  = {abs/2512.13564},
  year    = {2025}
}

@article{param,
  author  = {Song Wang and
             Yaochen Zhu and
             Haochen Liu and
             Zaiyi Zheng and
             Chen Chen and
             Jundong Li},
  title   = {Knowledge Editing for Large Language Models: {A} Survey},
  journal = {{ACM} Comput. Surv.},
  volume  = {57},
  number  = {3},
  pages   = {59:1--59:37},
  year    = {2025}
}

@article{nguyen2010bridging,
  title   = {Bridging semantic gaps in information retrieval: Context-based approaches},
  author  = {Nguyen, Cam-Tu},
  journal = {ACM VLDB},
  volume  = {10},
  year    = {2010}
}

@inproceedings{LongMemEval,
  author       = {Di Wu and
                  Hongwei Wang and
                  Wenhao Yu and
                  Yuwei Zhang and
                  Kai{-}Wei Chang and
                  Dong Yu},
  title        = {LongMemEval: Benchmarking Chat Assistants on Long-Term Interactive
                  Memory},
  booktitle    = {{ICLR}},
  publisher    = {OpenReview.net},
  year         = {2025}
}

@inbook{JohnsonLaird2008,
  place     = {Cambridge},
  title     = {Mental Models and Deductive Reasoning},
  booktitle = {Reasoning: Studies of Human Inference and its Foundations},
  publisher = {Cambridge University Press},
  author    = {Johnson-Laird, Philip N.},
  editor    = {Adler, Jonathan E. and Rips, Lance J.Editors},
  year      = {2008},
  pages     = {206–222}
}

@inproceedings{reasoningonfact,
  author    = {Harsh Trivedi and
               Niranjan Balasubramanian and
               Tushar Khot and
               Ashish Sabharwal},
  title     = {Interleaving Retrieval with Chain-of-Thought Reasoning for Knowledge-Intensive
               Multi-Step Questions},
  booktitle = {{ACL} {(1)}},
  pages     = {10014--10037},
  publisher = {Association for Computational Linguistics},
  year      = {2023}
}

@article{retllm,
  author  = {Ali Modarressi and
             Ayyoob Imani and
             Mohsen Fayyaz and
             Hinrich Sch{\"{u}}tze},
  title   = {{RET-LLM:} Towards a General Read-Write Memory for Large Language
             Models},
  journal = {CoRR},
  volume  = {abs/2305.14322},
  year    = {2023}
}

@article{packer2023memgpt,
  author  = {Charles Packer and
             Vivian Fang and
             Shishir G. Patil and
             Kevin Lin and
             Sarah Wooders and
             Joseph E. Gonzalez},
  title   = {MemGPT: Towards LLMs as Operating Systems},
  journal = {CoRR},
  volume  = {abs/2310.08560},
  year    = {2023}
}

@inproceedings{SCM,
  author       = {Bing Wang and
                  Xinnian Liang and
                  Jian Yang and
                  Hui Huang and
                  Zhenhe Wu and
                  Shuangzhi Wu and
                  Zejun Ma and
                  Zhoujun Li},
  title        = {{SCM:} Enhancing Large Language Model with Self-Controlled Memory
                  Framework},
  booktitle    = {{DASFAA} {(6)}},
  series       = {Lecture Notes in Computer Science},
  pages        = {188--203},
  publisher    = {Springer},
  year         = {2025}
}

@article{mem0,
  author    = {Prateek Chhikara and
               Dev Khant and
               Saket Aryan and
               Taranjeet Singh and
               Deshraj Yadav},
  title     = {Mem0: Building Production-Ready {AI} Agents with Scalable Long-Term
               Memory},
  booktitle = {{ECAI}},
  series    = {Frontiers in Artificial Intelligence and Applications},
  pages     = {2993--3000},
  publisher = {{IOS} Press},
  year      = {2025}
}

@inproceedings{Zhong2024MemoryBank,
  author    = {Wanjun Zhong and
               Lianghong Guo and
               Qiqi Gao and
               He Ye and
               Yanlin Wang},
  title     = {MemoryBank: Enhancing Large Language Models with Long-Term Memory},
  booktitle = {{AAAI}},
  pages     = {19724--19731},
  publisher = {{AAAI} Press},
  year      = {2024}
}

@article{zep,
  author  = {Preston Rasmussen and
             Pavlo Paliychuk and
             Travis Beauvais and
             Jack Ryan and
             Daniel Chalef},
  title   = {Zep: {A} Temporal Knowledge Graph Architecture for Agent Memory},
  journal = {CoRR},
  volume  = {abs/2501.13956},
  year    = {2025}
}

@inproceedings{Amem,
  author  = {Wujiang Xu and
             Zujie Liang and
             Kai Mei and
             Hang Gao and
             Juntao Tan and
             Yongfeng Zhang},
  title   = {{A-MEM:} Agentic Memory for {LLM} Agents},
  journal = {CoRR},
  volume  = {abs/2502.12110},
  year    = {2025}
}

@inproceedings{meminsight,
  author    = {Rana Salama and
               Jason Cai and
               Michelle Yuan and
               Anna Currey and
               Monica Sunkara and
               Yi Zhang and
               Yassine Benajiba},
  title     = {MemInsight: Autonomous Memory Augmentation for {LLM} Agents},
  booktitle = {{EMNLP}},
  pages     = {33136--33152},
  publisher = {Association for Computational Linguistics},
  year      = {2025}
}

@inproceedings{Irrelevant,
  author    = {Ori Yoran and
               Tomer Wolfson and
               Ori Ram and
               Jonathan Berant},
  title     = {Making Retrieval-Augmented Language Models Robust to Irrelevant Context},
  booktitle = {{ICLR}},
  publisher = {OpenReview.net},
  year      = {2024}
}

@inproceedings{longintput,
  author    = {Bowen Jin and
               Jinsung Yoon and
               Jiawei Han and
               Sercan {\"{O}}. Arik},
  title     = {Long-Context LLMs Meet {RAG:} Overcoming Challenges for Long Inputs
               in {RAG}},
  booktitle = {{ICLR}},
  publisher = {OpenReview.net},
  year      = {2025}
}

@article{erickson1998rules,
  title     = {Rules and exemplars in category learning.},
  author    = {Erickson, Michael A and Kruschke, John K},
  journal   = {Journal of Experimental Psychology: General},
  volume    = {127},
  number    = {2},
  pages     = {107},
  year      = {1998},
  publisher = {American Psychological Association}
}

@article{HGMEM,
  author       = {Chulun Zhou and
                  Chunkang Zhang and
                  Guoxin Yu and
                  Fandong Meng and
                  Jie Zhou and
                  Wai Lam and
                  Mo Yu},
  title        = {Improving Multi-step {RAG} with Hypergraph-based Memory for Long-Context
                  Complex Relational Modeling},
  journal      = {CoRR},
  volume       = {abs/2512.23959},
  year         = {2025}
}

@inproceedings{Workflow,
  author    = {Zora Zhiruo Wang and
               Jiayuan Mao and
               Daniel Fried and
               Graham Neubig},
  title     = {Agent Workflow Memory},
  booktitle = {{ICML}},
  series    = {Proceedings of Machine Learning Research},
  publisher = {{PMLR} / OpenReview.net},
  year      = {2025}
}

@inproceedings{ExpeL,
  author    = {Andrew Zhao and
               Daniel Huang and
               Quentin Xu and
               Matthieu Lin and
               Yong{-}Jin Liu and
               Gao Huang},
  title     = {ExpeL: {LLM} Agents Are Experiential Learners},
  booktitle = {{AAAI}},
  pages     = {19632--19642},
  publisher = {{AAAI} Press},
  year      = {2024}
}

@inproceedings{shinn2024reflexion,
  author    = {Noah Shinn and
               Federico Cassano and
               Ashwin Gopinath and
               Karthik Narasimhan and
               Shunyu Yao},
  title     = {Reflexion: language agents with verbal reinforcement learning},
  booktitle = {NeurIPS},
  year      = {2023}
}

@article{horn1951logic,
  author  = {Alfred Horn},
  title   = {On Sentences Which are True of Direct Unions of Algebras},
  journal = {J. Symb. Log.},
  volume  = {16},
  number  = {1},
  pages   = {14--21},
  year    = {1951}
}

@inproceedings{timelinememory,
  author    = {Kai Tzu{-}iunn Ong and
               Namyoung Kim and
               Minju Gwak and
               Hyungjoo Chae and
               Taeyoon Kwon and
               Yohan Jo and
               Seung{-}won Hwang and
               Dongha Lee and
               Jinyoung Yeo},
  title     = {Towards Lifelong Dialogue Agents via Timeline-based Memory Management},
  booktitle = {{NAACL} (Long Papers)},
  pages     = {8631--8661},
  publisher = {Association for Computational Linguistics},
  year      = {2025}
}

@article{RGMem,
  author  = {Ao Tian and
             Yunfeng Lu and
             Xinxin Fan and
             Changhao Wang and
             Lanzhi Zhou and
             Yeyao Zhang and
             Yanfang Liu},
  title   = {RGMem: Renormalization Group-based Memory Evolution for Language Agent
             User Profile},
  journal = {CoRR},
  volume  = {abs/2510.16392},
  year    = {2025}
}

@article{MemoryR1,
  author  = {Sikuan Yan and
             Xiufeng Yang and
             Zuchao Huang and
             Ercong Nie and
             Zifeng Ding and
             Zonggen Li and
             Xiaowen Ma and
             Hinrich Sch{\"{u}}tze and
             Volker Tresp and
             Yunpu Ma},
  title   = {Memory-R1: Enhancing Large Language Model Agents to Manage and Utilize
             Memories via Reinforcement Learning},
  journal = {CoRR},
  volume  = {abs/2508.19828},
  year    = {2025}
}

@article{Probabilistic,
  author  = {Yoshua Bengio and
             R{\'{e}}jean Ducharme and
             Pascal Vincent and
             Christian Janvin},
  title   = {A Neural Probabilistic Language Model},
  journal = {J. Mach. Learn. Res.},
  volume  = {3},
  pages   = {1137--1155},
  year    = {2003}
}

@inproceedings{sun2024tog,
  author    = {Jiashuo Sun and
               Chengjin Xu and
               Lumingyuan Tang and
               Saizhuo Wang and
               Chen Lin and
               Yeyun Gong and
               Lionel M. Ni and
               Heung{-}Yeung Shum and
               Jian Guo},
  title     = {Think-on-Graph: Deep and Responsible Reasoning of Large Language Model
               on Knowledge Graph},
  booktitle = {{ICLR}},
  publisher = {OpenReview.net},
  year      = {2024}
}

@article{wang2023voyager,
  author  = {Guanzhi Wang and
             Yuqi Xie and
             Yunfan Jiang and
             Ajay Mandlekar and
             Chaowei Xiao and
             Yuke Zhu and
             Linxi Fan and
             Anima Anandkumar},
  title   = {Voyager: An Open-Ended Embodied Agent with Large Language Models},
  journal = {Trans. Mach. Learn. Res.},
  volume  = {2024},
  year    = {2024}
}

@inproceedings{yao2022react,
  author    = {Shunyu Yao and
               Jeffrey Zhao and
               Dian Yu and
               Nan Du and
               Izhak Shafran and
               Karthik R. Narasimhan and
               Yuan Cao},
  title     = {ReAct: Synergizing Reasoning and Acting in Language Models},
  booktitle = {{ICLR}},
  publisher = {OpenReview.net},
  year      = {2023}
}

@inproceedings{yuan2025metakgrag,
  author    = {Xujie Yuan and
               Shimin Di and
               Jielong Tang and
               Libin Zheng and
               Jian Yin},
  title     = {Towards Self-cognitive Exploration: Metacognitive Knowledge Graph
               Retrieval Augmented Generation},
  booktitle = {{KDD} {(1)}},
  pages     = {1844--1855},
  publisher = {{ACM}},
  year      = {2026}
}

@article{guu2020realm,
  author       = {Kelvin Guu and
                  Kenton Lee and
                  Zora Tung and
                  Panupong Pasupat and
                  Ming{-}Wei Chang},
  title        = {{REALM:} Retrieval-Augmented Language Model Pre-Training},
  journal      = {CoRR},
  volume       = {abs/2002.08909},
  year         = {2020}
}

@inproceedings{lewis2020rag,
  author       = {Patrick Lewis and
                  Ethan Perez and
                  Aleksandra Piktus and
                  Fabio Petroni and
                  Vladimir Karpukhin and
                  Naman Goyal and
                  Heinrich K{\"{u}}ttler and
                  Mike Lewis and
                  Wen{-}tau Yih and
                  Tim Rockt{\"{a}}schel and
                  Sebastian Riedel and
                  Douwe Kiela},
  title        = {Retrieval-Augmented Generation for Knowledge-Intensive {NLP} Tasks},
  booktitle    = {NeurIPS},
  year         = {2020}
}

@inproceedings{borgeaud2022retro,
  author       = {Sebastian Borgeaud and
                  Arthur Mensch and
                  Jordan Hoffmann and
                  Trevor Cai and
                  Eliza Rutherford and
                  Katie Millican and
                  George van den Driessche and
                  Jean{-}Baptiste Lespiau and
                  Bogdan Damoc and
                  Aidan Clark and
                  Diego de Las Casas and
                  Aurelia Guy and
                  Jacob Menick and
                  Roman Ring and
                  Tom Hennigan and
                  Saffron Huang and
                  Loren Maggiore and
                  Chris Jones and
                  Albin Cassirer and
                  Andy Brock and
                  Michela Paganini and
                  Geoffrey Irving and
                  Oriol Vinyals and
                  Simon Osindero and
                  Karen Simonyan and
                  Jack W. Rae and
                  Erich Elsen and
                  Laurent Sifre},
  title        = {Improving Language Models by Retrieving from Trillions of Tokens},
  booktitle    = {{ICML}},
  series       = {Proceedings of Machine Learning Research},
  pages        = {2206--2240},
  publisher    = {{PMLR}},
  year         = {2022}
}

@inproceedings{asai2024selfrag,
  author       = {Akari Asai and
                  Zeqiu Wu and
                  Yizhong Wang and
                  Avirup Sil and
                  Hannaneh Hajishirzi},
  title        = {Self-RAG: Learning to Retrieve, Generate, and Critique through Self-Reflection},
  booktitle    = {{ICLR}},
  publisher    = {OpenReview.net},
  year         = {2024}
}

@inproceedings{sarthi2024raptor,
  author       = {Parth Sarthi and
                  Salman Abdullah and
                  Aditi Tuli and
                  Shubh Khanna and
                  Anna Goldie and
                  Christopher D. Manning},
  title        = {{RAPTOR:} Recursive Abstractive Processing for Tree-Organized Retrieval},
  booktitle    = {{ICLR}},
  publisher    = {OpenReview.net},
  year         = {2024}
}

@inproceedings{gutierrez2024hipporag,
  author       = {Bernal Jimenez Gutierrez and
                  Yiheng Shu and
                  Yu Gu and
                  Michihiro Yasunaga and
                  Yu Su},
  title        = {HippoRAG: Neurobiologically Inspired Long-Term Memory for Large Language
                  Models},
  booktitle    = {NeurIPS},
  year         = {2024}
}

@inproceedings{he2024gretriever,
  author       = {Xiaoxin He and
                  Yijun Tian and
                  Yifei Sun and
                  Nitesh V. Chawla and
                  Thomas Laurent and
                  Yann LeCun and
                  Xavier Bresson and
                  Bryan Hooi},
  title        = {G-Retriever: Retrieval-Augmented Generation for Textual Graph Understanding
                  and Question Answering},
  booktitle    = {NeurIPS},
  year         = {2024}
}

@inproceedings{li2024dalk,
  author       = {Dawei Li and
                  Shu Yang and
                  Zhen Tan and
                  Jae Young Baik and
                  Sukwon Yun and
                  Joseph Lee and
                  Aaron Chacko and
                  Bojian Hou and
                  Duy Duong{-}Tran and
                  Ying Ding and
                  Huan Liu and
                  Li Shen and
                  Tianlong Chen},
  title        = {{DALK:} Dynamic Co-Augmentation of LLMs and {KG} to answer Alzheimer's
                  Disease Questions with Scientific Literature},
  booktitle    = {{EMNLP} (Findings)},
  series       = {Findings of {ACL}},
  pages        = {2187--2205},
  publisher    = {Association for Computational Linguistics},
  year         = {2024}
}

@article{edge2024graphrag,
  author       = {Darren Edge and
                  Ha Trinh and
                  Newman Cheng and
                  Joshua Bradley and
                  Alex Chao and
                  Apurva Mody and
                  Steven Truitt and
                  Jonathan Larson},
  title        = {From Local to Global: {A} Graph {RAG} Approach to Query-Focused Summarization},
  journal      = {CoRR},
  volume       = {abs/2404.16130},
  year         = {2024}
}

@inproceedings{guo2024lightrag,
  author       = {Zirui Guo and
                  Lianghao Xia and
                  Yanhua Yu and
                  Tu Ao and
                  Chao Huang},
  title        = {LightRAG: Simple and Fast Retrieval-Augmented Generation},
  booktitle    = {{EMNLP} (Findings)},
  pages        = {10746--10761},
  publisher    = {Association for Computational Linguistics},
  year         = {2025}
}

@inproceedings{DA-RAG,
  author       = {Xingyuan Zeng and
                  Zuohan Wu and
                  Yue Wang and
                  Chen Zhang and
                  Quanming Yao and
                  Libin Zheng and
                  Jian Yin},
  title        = {{DA-RAG:} Dynamic Attributed Community Search for Retrieval-Augmented
                  Generation},
  booktitle    = {{WWW}},
  pages        = {2195--2206},
  publisher    = {{ACM}},
  year         = {2026}
}

@article{communityRAG,
  author       = {Rong{-}Ching Chang and
                  Jiawei Zhang},
  title        = {CommunityKG-RAG: Leveraging Community Structures in Knowledge Graphs for Advanced Retrieval-Augmented Generation in Fact-Checking},
  journal      = {CoRR},
  volume       = {abs/2408.08535},
  year         = {2024}
}

@inproceedings{BM25,
  author       = {Stephen E. Robertson and
                  Steve Walker and
                  Susan Jones and
                  Micheline Hancock{-}Beaulieu and
                  Mike Gatford},
  title        = {Okapi at {TREC-3}},
  booktitle    = {{TREC}},
  series       = {{NIST} Special Publication},
  volume       = {500-225},
  pages        = {109--126},
  publisher    = {National Institute of Standards and Technology {(NIST)}},
  year         = {1994}
}

@article{lossinmiddle,
  author       = {Nelson F. Liu and
                  Kevin Lin and
                  John Hewitt and
                  Ashwin Paranjape and
                  Michele Bevilacqua and
                  Fabio Petroni and
                  Percy Liang},
  title        = {Lost in the Middle: How Language Models Use Long Contexts},
  journal      = {Trans. Assoc. Comput. Linguistics},
  volume       = {12},
  pages        = {157--173},
  year         = {2024}
}

@article{hallucination2,
author = {Ji, Ziwei and Lee, Nayeon and Frieske, Rita and Yu, Tiezheng and Su, Dan and Xu, Yan and Ishii, Etsuko and Bang, Ye Jin and Madotto, Andrea and Fung, Pascale},
title = {Survey of Hallucination in Natural Language Generation},
year = {2023},
issue_date = {December 2023},
publisher = {Association for Computing Machinery},
address = {New York, NY, USA},
volume = {55},
number = {12},
issn = {0360-0300},
url = {https://doi.org/10.1145/3571730},
doi = {10.1145/3571730},
journal = {ACM Comput. Surv.},
month = mar,
articleno = {248},
numpages = {38}
}

@article{hallucination,
  author       = {Lei Huang and
                  Weijiang Yu and
                  Weitao Ma and
                  Weihong Zhong and
                  Zhangyin Feng and
                  Haotian Wang and
                  Qianglong Chen and
                  Weihua Peng and
                  Xiaocheng Feng and
                  Bing Qin and
                  Ting Liu},
  title        = {A Survey on Hallucination in Large Language Models: Principles, Taxonomy,
                  Challenges, and Open Questions},
  journal      = {CoRR},
  volume       = {abs/2311.05232},
  year         = {2023}
}

@inproceedings{rag-sur2,
  author       = {Wenqi Fan and
                  Yujuan Ding and
                  Liangbo Ning and
                  Shijie Wang and
                  Hengyun Li and
                  Dawei Yin and
                  Tat{-}Seng Chua and
                  Qing Li},
  title        = {A Survey on {RAG} Meeting LLMs: Towards Retrieval-Augmented Large
                  Language Models},
  booktitle    = {{KDD}},
  pages        = {6491--6501},
  publisher    = {{ACM}},
  year         = {2024}
}

@article{rag-survey,
  author       = {Yunfan Gao and
                  Yun Xiong and
                  Xinyu Gao and
                  Kangxiang Jia and
                  Jinliu Pan and
                  Yuxi Bi and
                  Yi Dai and
                  Jiawei Sun and
                  Qianyu Guo and
                  Meng Wang and
                  Haofen Wang},
  title        = {Retrieval-Augmented Generation for Large Language Models: {A} Survey},
  journal      = {CoRR},
  volume       = {abs/2312.10997},
  year         = {2023}
}

@article{lightRaG,
  author       = {Zirui Guo and
                  Lianghao Xia and
                  Yanhua Yu and
                  Tu Ao and
                  Chao Huang},
  title        = {LightRAG: Simple and Fast Retrieval-Augmented Generation},
  journal      = {CoRR},
  volume       = {abs/2410.05779},
  year         = {2024}
}

@article{GraphRAG,
  author       = {Darren Edge and
                  Ha Trinh and
                  Newman Cheng and
                  Joshua Bradley and
                  Alex Chao and
                  Apurva Mody and
                  Steven Truitt and
                  Jonathan Larson},
  title        = {From Local to Global: {A} Graph {RAG} Approach to Query-Focused Summarization},
  journal      = {CoRR},
  volume       = {abs/2404.16130},
  year         = {2024}
}

@article{ArchRAG,
  author       = {Shu Wang and
                  Yixiang Fang and
                  Yingli Zhou and
                  Xilin Liu and
                  Yuchi Ma},
  title        = {ArchRAG: Attributed Community-based Hierarchical Retrieval-Augmented
                  Generation},
  journal      = {CoRR},
  volume       = {abs/2502.09891},
  year         = {2025}
}


\end{document}